\documentclass[journal]{IEEEtran}
\usepackage{amssymb}
\usepackage{amsfonts}
\usepackage{booktabs}
\usepackage{multirow}
\usepackage{graphicx}
\usepackage{subcaption}
\usepackage{amsmath}
\ifCLASSINFOpdf
\else
   \usepackage[dvips]{graphicx}
\fi
\usepackage{url}

\begin{document}

\title{LiquidTAD: Efficient Temporal Action Detection via Parallel Liquid-Inspired Temporal Relaxation}

\author{
    Zepeng Sun\thanks{$^\ast$These authors contributed equally to this work.},
    Naichuan Zheng$^{\ast}$,
    Hailun Xia\thanks{Corresponding author. Email: xiahailun@bupt.edu.cn},
    Junjie Wu,
    Liwei Bao,
    Xiaotai Zhang
}

\markboth{Journal of \LaTeX\ Class Files, Vol. 14, No. 8, August 2015}
{Shell \MakeLowercase{\textit{et al.}}: Bare Demo of IEEEtran.cls for IEEE Journals}
\maketitle

\begin{abstract}
Temporal Action Detection (TAD) requires precise localization of action boundaries within long, untrimmed video sequences. While current high-performing methods achieve strong accuracy, they are often characterized by excessive parameter counts, substantial computational overhead, and a reliance on specialized operators that hinder deployment across diverse hardware platforms. This paper presents \textbf{LiquidTAD}, a framework that distills the exponential relaxation prior of liquid neural dynamics into a parallel temporal operator, rather than reproducing full Liquid Neural Network (LNN) dynamics. By introducing a \textbf{Parallel Liquid-inspired Relaxation} mechanism, sequential ODE solving is avoided through a fully vectorized, non-recursive formulation built entirely upon standard neural operations, enabling hardware-agnostic deployment with linear complexity with respect to the temporal length. A complementary \textbf{Hierarchical Decay-Rate Sharing Strategy} further adapts this relaxation prior across feature pyramid levels~\cite{lin2017fpn}, stabilizing optimization and implicitly compensating for temporal compression in deeper layers. Experimental evaluations on THUMOS-14~\cite{idrees2017thumos} and ActivityNet-1.3~\cite{caba2015activitynet} demonstrate that LiquidTAD achieves accuracy competitive with strong baselines while substantially lowering the model footprint. Specifically, on THUMOS-14, LiquidTAD achieves \textbf{69.46\% average mAP} with only \textbf{10.82M parameters} and \textbf{27.17G FLOPs}, reducing the parameter count by over \textbf{60\%} compared with ActionFormer~\cite{zhang2022actionformer}.
\end{abstract}


\IEEEpeerreviewmaketitle

\section{Introduction}

\begin{figure}[t]
  \centering
  \includegraphics[width=\columnwidth]{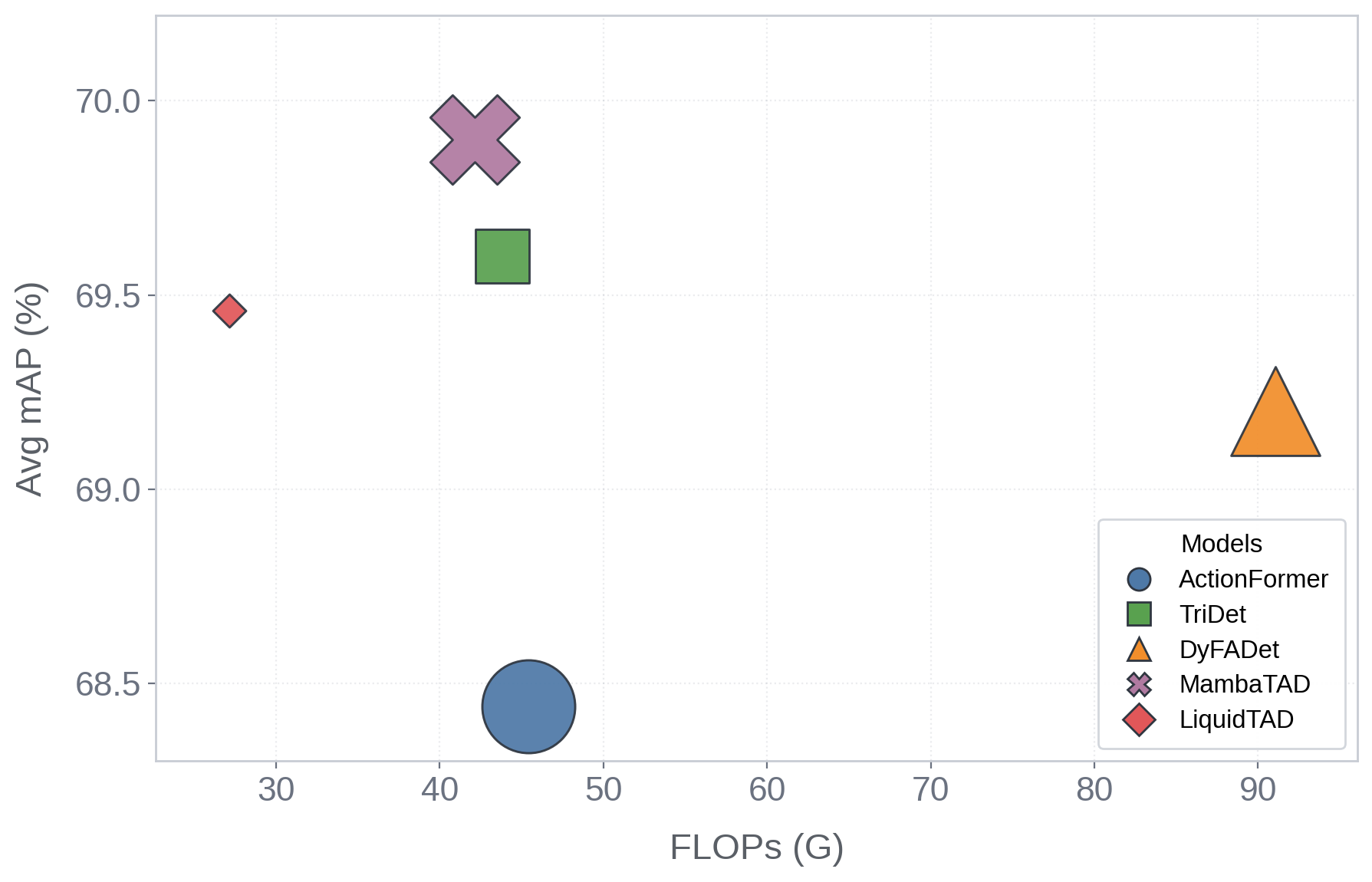}
  \caption{Performance versus parameter complexity on THUMOS-14. The horizontal axis shows FLOPs, the vertical axis shows average mAP, and the size of each point represents the model parameter count. LiquidTAD achieves competitive mAP with a substantially smaller detection head, highlighting its efficiency-accuracy trade-off.}
  \label{fig:pareto}
\end{figure}

Temporal Action Detection (TAD) stands as a pivotal task in video analysis, requiring the precise localization of action boundaries within long, untrimmed video sequences. To balance computational efficiency and detection accuracy, a widely adopted approach relies on processing pre-extracted video features through a temporal modeling head. In the pursuit of higher detection accuracy, Transformer-based architectures have emerged as the dominant paradigm, leveraging self-attention mechanisms to capture long-range temporal dependencies. However, the performance of these models often comes at the cost of excessive parameter counts and heavy computational overhead, frequently necessitating specialized acceleration operators to manage the complexity of self-attention. Representative works such as ActionFormer~\cite{zhang2022actionformer} and TadTR~\cite{liu2022tadtr} exemplify this trend. While the recent introduction of State Space Models (SSMs) -- initially formalized through structured state space models~\cite{gu2022s4} and further popularized by Mamba~\cite{gu2023mamba} -- has alleviated the parameter count and computational costs to a certain extent, these methods still largely rely on specialized operators and custom CUDA kernels to maintain competitive efficiency. This reliance complicates deployment across diverse hardware platforms, particularly where standard library support is preferred.

Beyond the reliance on specialized operators, a fundamental limitation of prevailing TAD methodologies lies in their modeling paradigm for temporal progression. While current architectures effectively aggregate context across discrete tokens, they typically lack an explicit, lightweight mechanism to model temporal persistence and decay patterns in sequential video data. Although traditional recurrent architectures attempt to capture such temporal dynamics, they are constrained by their step-by-step processing nature. This serial bottleneck limits hardware-level vectorization, making them costly for processing long, untrimmed video sequences. In this context, concepts derived from Liquid Neural Networks (LNNs)---specifically the Liquid Time-Constant (LTC) networks~\cite{hasani2021ltc} and Closed-form Continuous-time (CfC) networks~\cite{hasani2022cfc}---offer a compelling inductive bias through their temporal decay mechanism governed by learned time constants. By modulating state transitions through an exponential decay factor, this approach provides a structurally elegant representation of temporal evolution. However, traditional LNN implementations share the same limitation as standard recurrent networks: they rely on sequential solvers (e.g., ODE solvers~\cite{chen2018neuralode}) that hinder parallelization. The critical challenge is therefore to decouple this effective exponential relaxation prior from restrictive sequential processing, reformulating it into a fully parallelizable operator that enables efficient computation using standard neural operations.

To address these limitations, we introduce \textbf{LiquidTAD}, an efficient TAD framework that distills the exponential relaxation prior of liquid neural dynamics into a non-recursive parallel temporal operator, rather than reproducing full LNN dynamics. Specifically, LiquidTAD proposes a \textbf{Parallel Liquid-inspired Relaxation} mechanism. By reformulating the relaxation prior as a vectorized feature update parameterized by a learned decay rate ($\lambda$) and a fixed structural time step ($\Delta t$), this mechanism computes temporal representations entirely in parallel along the sequence dimension using standard neural operations. Consequently, it avoids the serial bottlenecks of traditional ODE solvers while remaining hardware-agnostic across diverse deployment platforms.

Furthermore, to effectively adapt this temporal modeling mechanism across the multi-scale architecture of the TAD framework, LiquidTAD employs a \textbf{Hierarchical Decay-Rate Sharing Strategy}. Rather than utilizing per-channel decay rates---which can lead to redundant temporal dynamics and suboptimal optimization---the model learns an independent, channel-shared scalar decay rate for each block within the feature pyramid~\cite{lin2017fpn}. This strategy encourages the network to capture a unified, scale-specific temporal decay, yielding more stable representations and improved detection performance. Extensive experiments across THUMOS-14~\cite{idrees2017thumos} and ActivityNet-1.3~\cite{caba2015activitynet} benchmarks demonstrate that LiquidTAD achieves competitive detection accuracy while substantially reducing total parameters and FLOPs compared to Transformer-based and CNN-based baselines. As illustrated in Figure~\ref{fig:pareto}, LiquidTAD establishes a strong efficiency-accuracy Pareto front. Owing to its standard-operator-driven design, LiquidTAD also exhibits low inference latency on standard CPUs, significantly outperforming both attention-based and SSM-based architectures in deployment-constrained scenarios.

The primary contributions of this work are summarized as follows:
\begin{itemize}
    \item To the best of our knowledge, \textbf{LiquidTAD} is the first TAD framework to exploit a liquid-inspired exponential relaxation prior for efficient temporal modeling, striking a favorable balance between temporal modeling capability and computational efficiency.
    \item We propose a \textbf{Parallel Liquid-inspired Relaxation} mechanism. By reformulating the exponential relaxation prior as a fully vectorized, non-recursive feature update, we avoid the serial bottlenecks of traditional ODE solvers~\cite{chen2018neuralode}, enabling hardware-level parallelism using standard neural operations.
    \item We design a \textbf{Hierarchical Decay-Rate Sharing Strategy} tailored for multi-scale feature pyramids~\cite{lin2017fpn}. By constraining the decay rate to a single learnable scalar per block, this mechanism stabilizes optimization and allows the network to learn scale-aligned temporal dynamics without introducing redundant per-channel parameters.
    \item Extensive experiments on THUMOS-14~\cite{idrees2017thumos} and ActivityNet-1.3~\cite{caba2015activitynet} demonstrate the effectiveness of our approach. LiquidTAD achieves competitive detection accuracy with substantially reduced model complexity, while demonstrating large efficiency gains over sequential liquid backends and low CPU inference latency compared with attention-based and SSM-based models.
\end{itemize}

\section{Related Work}

\subsection{Temporal Action Detection}
Temporal Action Detection (TAD) has undergone substantial evolution. A comprehensive survey and standardized comparison of representative methods can be found in OpenTAD~\cite{opentad2024}, which provides a unified re-implementation framework across diverse TAD paradigms.

\textbf{Early proposal-based and segment-based methods.} The field was significantly shaped by approaches that decompose the detection problem into temporal proposal generation followed by classification. Structured Segment Networks (SSN)~\cite{zhao2017ssn} introduced activity completeness modeling via structured temporal segments, while BMN~\cite{lin2019bmn} proposed a boundary-matching scheme for high-quality proposal generation. BSN~\cite{lin2018bsn} further improved boundary sensitivity through a local-to-global proposal generation pipeline. DBG~\cite{lin2020dbg} accelerated proposal generation via a dense boundary approach. These methods established strong baselines but require separate proposal and classification stages, limiting end-to-end optimization.

\textbf{Anchor-free and one-stage detectors.} To overcome the limitations of two-stage pipelines, anchor-free methods have been proposed. AFSD~\cite{lin2021afsd} introduced the first one-stage, anchor-free TAD framework with salient boundary cues. RTD-Net~\cite{tan2021rtdnet} employed relaxed transformer decoders for direct proposal generation without anchor boxes. TCANet~\cite{qing2021tcanet} further refined temporal proposals via context-aware aggregation. PointTAD~\cite{tan2022pointtad} extended TAD to multi-label scenarios through learnable query points, enabling detection of concurrent and overlapping actions.

\textbf{Transformer-based models}, such as ActionFormer~\cite{zhang2022actionformer} and TadTR~\cite{liu2022tadtr}, demonstrated the power of self-attention for global temporal context aggregation, but they often incur heavy FLOPs and memory consumption as sequence length grows. TALLFormer~\cite{cheng2022tallformer} addressed very long videos via a memory-augmented transformer that combines short- and long-range temporal modeling. To improve efficiency, \textbf{CNN-based approaches} like TriDet~\cite{shi2023tridet} employ 1D temporal convolutions for localized context aggregation; however, they are constrained by fixed receptive fields and typically require stacking numerous layers to extend coverage, increasing the parameter footprint. More recent efficient temporal modeling methods, such as TemporalMaxer~\cite{tang2023temporalmaxer}, explore gated or convolutional temporal mixing to reduce computational overhead while maintaining competitive accuracy.

Recently, \textbf{State Space Models (SSMs)}~\cite{gu2022s4,gu2023mamba} have gained traction in TAD. MambaTAD~\cite{mambatad2024} adapts the Mamba architecture to balance performance and efficiency under a stronger detection framework. Although these methods achieve linear complexity in theory, their practical speedup relies on \textbf{hardware-specific custom kernels}. When deployed on general-purpose processors where these specialized kernels are unavailable, these models can experience sharp increases in inference latency. LiquidTAD addresses this limitation by building its relaxation-based temporal modeling using only standard neural operations, ensuring consistent efficiency across diverse hardware platforms without relying on custom CUDA kernels or specialized operators.

Although convolutional and gated temporal modeling methods improve efficiency, they typically do not explicitly parameterize temporal feature retention through a continuous-time-inspired decay rate. LiquidTAD differs by introducing an exponential relaxation prior that directly controls the balance between feature preservation and temporal stimulus injection, providing a physically interpretable inductive bias that is not explicitly modeled in standard convolutional or gated temporal operators.

\subsection{Liquid and Continuous-Time Neural Models}
Continuous-time neural models, including Liquid Neural Networks (LNNs), Liquid Time-Constant (LTC) networks~\cite{hasani2021ltc}, and Closed-form Continuous-time (CfC) networks~\cite{hasani2022cfc}, are designed to process sequential data by modeling state evolution through dynamic time constants ($\tau$). These models capture complex temporal dynamics via an exponential relaxation mechanism that governs how the network state transitions toward a target stimulus over time. This exponential relaxation prior constitutes a principled, physically interpretable inductive bias for temporal modeling.

Despite their theoretical appeal, traditional LNN and CfC implementations are typically realized through recurrent updates or sequential numerical integration via Neural ODE solvers~\cite{chen2018neuralode}. This reliance creates a strict serial processing bottleneck, leading to substantial computational overhead for the long, high-dimensional feature sequences typical in TAD. Critically, LiquidTAD does not attempt to reproduce these full recurrent dynamics. Instead, it identifies the exponential relaxation prior as the central inductive bias and reformulates it into a fully vectorized, non-recursive parallel operator. By decoupling the relaxation prior from sequential ODE solving and grounding it in pre-extracted video features, LiquidTAD preserves the continuous-time-inspired relaxation principle while achieving hardware-agnostic efficiency suitable for large-scale video understanding.

\section{Methodology}
\begin{figure*}[t]
  \centering
  \includegraphics[width=0.9\textwidth]{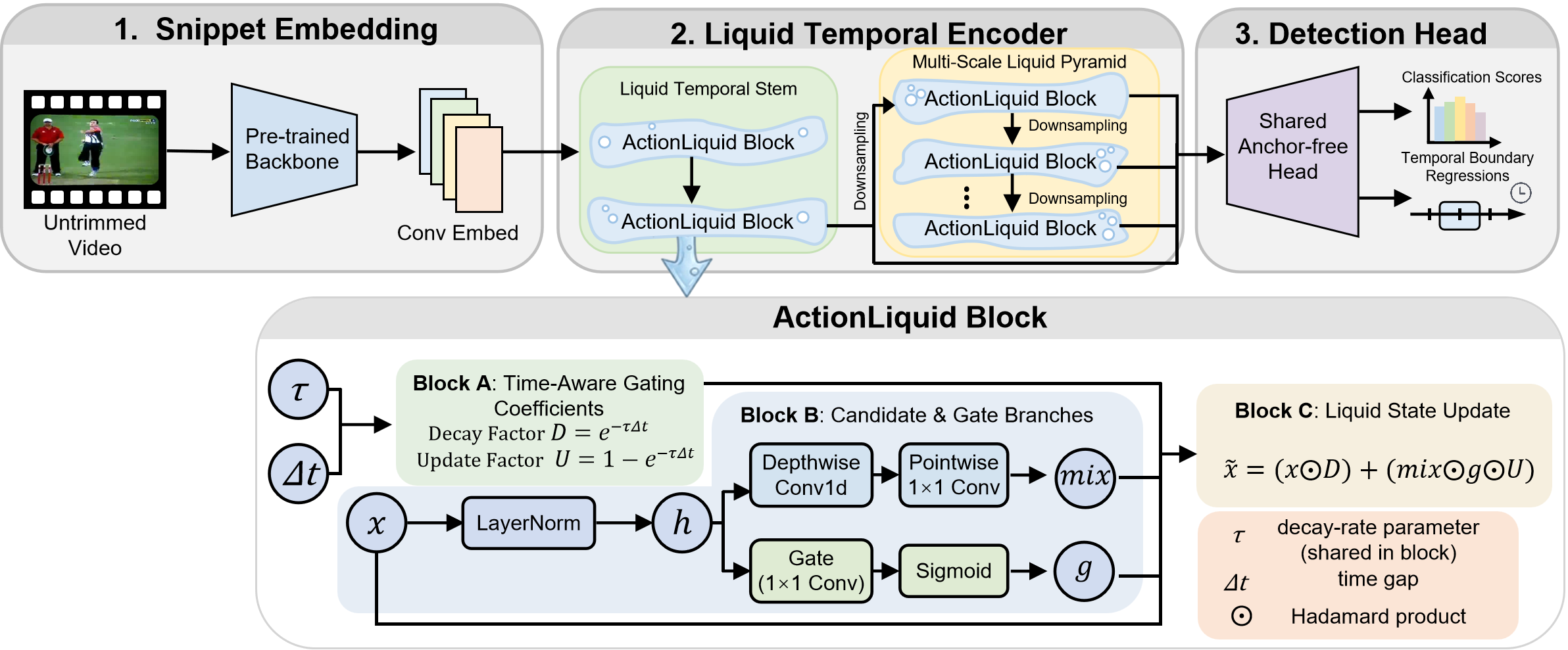}
  \caption{The overall architecture of LiquidTAD. The input video features are processed through a feature pyramid equipped with Liquid Temporal Stacks, utilizing Parallel Liquid-inspired Relaxation and a Hierarchical Decay-Rate Sharing Strategy to efficiently capture multi-scale temporal dynamics without relying on sequential ODE solvers.}
  \label{fig:architecture}
\end{figure*}
In this section, we present \textbf{LiquidTAD}, a framework that distills the continuous-time exponential relaxation prior of liquid dynamics into a parallelized operator for Temporal Action Detection (TAD). Our central design philosophy is as follows: rather than reproducing full LNN dynamics, we identify a key inductive bias they offer---exponential temporal relaxation---and reformulate it as a non-recursive operator suitable for hardware-level parallelism. We first provide an architectural overview, followed by the derivation of our parallel liquid-inspired relaxation and the hierarchical scale-adaptive parameterization strategy.

\subsection{Overall Architecture}
LiquidTAD follows the mainstream encode-and-detect paradigm. Given an untrimmed video sequence, a backbone (e.g., I3D~\cite{carreira2017i3d} or SlowFast~\cite{feichtenhofer2019slowfast}) extracts features $X \in \mathbb{R}^{T \times C}$. The core of our detector lies in the \textbf{Liquid Parallel Temporal Block (LPTB)}, which serves as the fundamental unit of our feature pyramid~\cite{lin2017fpn}.

As illustrated in Fig.~\ref{fig:architecture}, the input features pass through a multi-level feature pyramid where the temporal resolution is progressively downsampled. At each level $l$, an LPTB aggregates temporal context. Unlike traditional blocks that rely on heavy self-attention or deep 1D-CNNs, the LPTB implements the \textbf{Parallel Liquid-inspired Relaxation} (detailed in Sec. 3.2). This allows the model to capture scale-aware temporal decay with linear complexity while retaining a liquid-inspired exponential relaxation prior. Finally, classification and regression heads generate action proposals from the enriched multi-scale features.

\subsection{Parallel Liquid-inspired Relaxation}
The primary challenge in adopting Liquid Neural Networks (LNNs)~\cite{hasani2021ltc} for TAD is the fundamental conflict between their sequential ODE-based nature~\cite{chen2018neuralode} and the requirement for hardware-level parallelism. Critically, our goal is \emph{not} to approximate the full recurrent LNN dynamics, but to distill a key inductive bias---exponential temporal relaxation---into a non-recursive, fully vectorized operator. We achieve this through a four-step derivation that systematically identifies and eliminates each source of serial dependency.

\textbf{Step 1: Standard Liquid Relaxation Prior.}
Standard Closed-form Continuous (CfC) networks~\cite{hasani2022cfc} approximate the state evolution of a liquid system as:
\begin{equation}
x_{t+1} = e^{-\Delta t / \tau_t} \cdot x_t + \left(1 - e^{-\Delta t / \tau_t}\right) \cdot h(x_t, u_t)
\end{equation}
where $\tau_t$ is the dynamic time constant. This formulation encodes the exponential relaxation prior we seek to preserve: a physically interpretable continuous-time mechanism for temporal state decay. However, the recurrent dependency on $x_t$ and the input-dependent $\tau_t$ create severe serial bottlenecks that hinder efficient parallel computation. The remainder of our derivation is dedicated to removing these bottlenecks while retaining the relaxation prior itself.

\textbf{Step 2: Static Decay-Rate Stabilization.}
In TAD, input features $X_l$ are extracted by robust pre-trained backbones that already encode rich local and global temporal contexts. Computing an input-dependent time constant in this setting would introduce substantial sequential overhead, while providing limited additional benefit. We therefore replace the dynamic time constant with a block-wise learnable \textbf{decay rate} $\lambda_l$:
\begin{equation}
\lambda_l = \text{Softplus}(\rho_l) + \epsilon, \quad \alpha_l = \exp(-\lambda_l \Delta t)
\end{equation}
where $\lambda_l$ acts as the inverse time constant of the $l$-th temporal block (i.e., $\lambda_l = 1/\tau_l$), and $\alpha_l$ is the resulting retention coefficient. This makes the decay factor constant within each pyramid level, avoiding repeated input-dependent decay computation.

\textbf{Step 3: Reference-State Relaxation.}
In conventional recurrent liquid models, the hidden state $x_t$ recursively propagates historical information. However, in our feature-extraction paradigm, each token $X_{l,t}$ already provides a strong representation of its temporal window. Rather than maintaining an additional autoregressive hidden state---which would reintroduce sequential overhead---we use $X_{l,t}$ as the \textit{reference state} and relax it toward a learned temporal stimulus:
\begin{equation}
\text{out}_{l,t} = \alpha_l X_{l,t} + (1 - \alpha_l) \cdot S_\theta(X_l)_t
\end{equation}
This substitution is an intentional architectural choice, not a compromise. By grounding the relaxation in the already-rich pre-extracted features rather than a separately maintained hidden state, we render the computation fully parallelizable along the temporal dimension while retaining the exponential decay prior as a structural constraint on how information is blended. The resulting operator can therefore be viewed as a practical distillation of the LNN's exponential relaxation mechanism, decoupled from its recurrent constraint.

\textbf{Step 4: Parameterizing the Temporal Stimulus.}
We implement the stimulus $S_\theta(X_l)$ as a gated module parallelized across the sequence. To provide the stimulus with local temporal context, we use a depthwise convolution to aggregate neighboring features:
\begin{equation}
\begin{aligned}
\hat{X}_l &= \text{LayerNorm}(X_l) \\
\text{mix}_l &= \text{Dropout}(\text{Pointwise}(\text{Depthwise}(\hat{X}_l))) \\
g_l &= \sigma(\text{Gate}(\hat{X}_l))
\end{aligned}
\end{equation}
The final parallel update for the $l$-th level is formulated as:
\begin{equation}
\text{out}_l = \alpha_l X_l + (1 - \alpha_l) \cdot (\text{mix}_l \odot g_l)
\end{equation}
Intuitively, this formulation functions as a \textbf{time-scale adaptive relaxation filter}. The decay rate $\lambda_l$ explicitly controls the retention of the current feature state, while the gated stimulus $g_l \odot \text{mix}_l$ governs the injection of localized temporal information. Unlike standard residual connections, this formulation preserves an interpretable relaxation prior derived from continuous-time dynamics.

\subsection{Hierarchical Decay-Rate Sharing Strategy}
The parameterization of $\lambda_l$ is critical for maintaining stability across the feature pyramid~\cite{lin2017fpn}. As the sequence is downsampled, each token represents a progressively larger temporal window. Consequently, a fixed physical time-step $\Delta t$ becomes an unreliable measure of ``real-time'' in deeper layers.

To ensure scale-consistency, we further introduce a \textbf{Hierarchical Decay-Rate Sharing Strategy}. By constraining $\lambda_l$ to be a single learnable scalar shared across all channels within a specific LPTB, we encourage the network to learn a unified, level-aligned temporal decay rate. This strategy stabilizes the optimization against channel-wise noise and allows the model to implicitly compensate for the temporal compression of the pyramid, helping the learned relaxation dynamics better adapt to the varying durations of actions at different scales.

\section{Experiments}

\subsection{Experimental Setup}
We comprehensively evaluate LiquidTAD on two widely adopted benchmarks for Temporal Action Detection: THUMOS-14~\cite{idrees2017thumos} and ActivityNet-1.3~\cite{caba2015activitynet}. Following the standardized evaluation protocols of OpenTAD~\cite{opentad2024}, we report the mean Average Precision (mAP) at various Intersection over Union (IoU) thresholds. For fair comparisons, we extract features using standard backbones (I3D~\cite{carreira2017i3d} for THUMOS-14 and TSP~\cite{alwassel2021tsp} for ActivityNet-1.3) to ensure performance gains stem directly from our proposed temporal modeling backend.

\subsection{State-of-the-Art Comparison}

\textbf{Accuracy on Mainstream Benchmarks.} As shown in Table~\ref{tab:sota_thumos_anet}, LiquidTAD achieves competitive performance across both datasets while operating with substantially lower model complexity. On THUMOS-14, LiquidTAD achieves an average mAP of 69.46\%, comparable to TriDet~\cite{shi2023tridet} (69.60\%) and CausalTAD~\cite{causaltad2024} (69.75\%). On ActivityNet-1.3, LiquidTAD reaches 37.06\% average mAP, closely matching ActionFormer~\cite{zhang2022actionformer} (37.07\%). We note that MambaTAD~\cite{mambatad2024} is built upon a stronger detection framework (VideoMamba~\cite{li2024videomamba} + TriDet~\cite{shi2023tridet}), while LiquidTAD adopts the lighter ActionFormer framework. Under this more constrained setting, LiquidTAD still achieves comparable accuracy with significantly lower computational cost.

% THUMOS 和 ActivityNet 的 SOTA 对比表
\begin{table*}[t]
  \centering
  \caption{State-of-the-art comparison on THUMOS-14 and ActivityNet-1.3. Performance is measured by mAP at different IoU thresholds and average mAP. $\dagger$ denotes methods built upon a stronger detection framework.}
  \renewcommand{\arraystretch}{1.15}
  \resizebox{0.95\textwidth}{!}{
    \begin{tabular}{ll cccccc l cccc}
    \toprule
    \multirow{2}{*}{\textbf{Method}} & \multicolumn{7}{c}{\textbf{THUMOS-14}} & \multicolumn{5}{c}{\textbf{ActivityNet-1.3}} \\
    \cmidrule(lr){2-8} \cmidrule(lr){9-13}
    & \textbf{Backbone} & \textbf{0.3} & \textbf{0.4} & \textbf{0.5} & \textbf{0.6} & \textbf{0.7} & \textbf{Avg} & \textbf{Backbone} & \textbf{0.5} & \textbf{0.75} & \textbf{0.95} & \textbf{Avg} \\
    \midrule
    TSI~\cite{liu2021tsi}         & I3D & 62.56 & 57.00 & 50.22 & 40.18 & 30.17 & 48.03 & TSP & 52.44 & 35.57 & 9.80  & 35.36 \\
    G-TAD~\cite{xu2020gtad}       & I3D & 63.35 & 59.07 & 51.76 & 42.65 & 31.66 & 49.70 & TSP & 52.33 & 37.58 & 8.42  & 36.20 \\
    BMN~\cite{lin2019bmn}         & I3D & 64.99 & 60.70 & 54.54 & 44.11 & 34.16 & 51.70 & TSP & 52.90 & 37.30 & 9.67  & 36.40 \\
    TadTR~\cite{liu2022tadtr}     & I3D & 71.90 & 67.29 & 59.00 & 48.34 & 34.61 & 56.23 & TSP & 53.62 & 37.52 & 10.56 & 36.75 \\
    TemporalMaxer~\cite{tang2023temporalmaxer} & I3D & 82.80 & 78.90 & 71.80 & 60.50 & 44.70 & 67.75 & - & - & - & - & - \\
    ActionFormer~\cite{zhang2022actionformer}  & I3D & 83.78 & 80.06 & 73.16 & 60.46 & 44.72 & 68.44 & TSP & 55.08 & 38.27 & 8.91 & 37.07 \\
    DyFADet~\cite{dyfahdet2024}   & I3D & 84.00 & 80.10 & 72.70 & 61.10 & 47.90 & 69.20 & TSP+IV & 58.19 & 39.30 & 8.63 & 38.62 \\
    TriDet~\cite{shi2023tridet}   & I3D & 84.46 & 81.05 & 73.41 & 62.58 & 46.51 & 69.60 & TSP & 54.89 & 38.20 & 8.21 & 36.96 \\
    CausalTAD~\cite{causaltad2024}& I3D & 84.43 & 80.75 & 73.57 & 62.70 & 47.33 & 69.75 & TSP & 55.62 & 38.51 & 9.40 & 37.46 \\
    MambaTAD~\cite{mambatad2024} & I3D$^\dagger$ & 84.30 & 80.70 & 74.10 & 62.90 & 47.50 & 69.90 & TSP & 60.20 & 41.30 & 9.70 & 40.20 \\
    \midrule
    \textbf{LiquidTAD (Ours)} & I3D & 84.21 & 79.86 & 72.90 & 62.87 & 47.47 & 69.46 & TSP & 55.18 & 38.10 & 8.43 & 37.06 \\
    \bottomrule
    \end{tabular}%
  }
  \label{tab:sota_thumos_anet}
\end{table*}

\textbf{System-Level Efficiency.} LiquidTAD achieves these competitive results while substantially reducing the model footprint, as detailed in Table~\ref{tab:complexity}. All complexity figures are reported for the temporal detection head only, excluding the frozen feature extraction backbone. On THUMOS-14, LiquidTAD requires only 10.82M parameters and 27.17G FLOPs. Compared to ActionFormer~\cite{zhang2022actionformer} (29.25M parameters, 45.41G FLOPs), this represents a reduction of approximately 63\% in parameters and 40\% in FLOPs. Even compared to the CNN-based TriDet~\cite{shi2023tridet} (15.99M parameters) or the parameter-heavy CausalTAD~\cite{causaltad2024} (52.11M parameters), LiquidTAD demonstrates favorable computational efficiency while maintaining competitive detection accuracy.

% 复杂度对比表
\begin{table}[htbp]
  \centering
  \caption{Complexity comparison across datasets. Parameters (M) and FLOPs (G) are reported for the temporal detection head, excluding the feature extraction backbone. ``-'' indicates values not reported under a comparable counting protocol.}
  \renewcommand{\arraystretch}{1.1}
  \resizebox{0.8\columnwidth}{!}{
    \begin{tabular}{l cc cc}
    \toprule
    \multirow{2}{*}{\textbf{Method}} & \multicolumn{2}{c}{\textbf{THUMOS}} & \multicolumn{2}{c}{\textbf{ANet}} \\
    \cmidrule(lr){2-3} \cmidrule(lr){4-5}
    & \textbf{Param} & \textbf{FLOP} & \textbf{Param} & \textbf{FLOP} \\
    \midrule
    ActionFormer~\cite{zhang2022actionformer} & 29.25 & 45.41 & 6.94  & 3.48 \\
    TriDet~\cite{shi2023tridet}               & 15.99 & 43.84 & 12.81 & 10.10 \\
    CausalTAD~\cite{causaltad2024}            & 52.11 & -     & 12.75 & -    \\
    \midrule
    \textbf{LiquidTAD} & \textbf{10.82} & \textbf{27.17} & \textbf{2.32} & \textbf{1.96} \\
    \bottomrule
    \end{tabular}%
  }
  \label{tab:complexity}
\end{table}

\subsection{Qualitative Results}

\begin{figure}[htbp]
  \centering
  \includegraphics[width=\columnwidth]{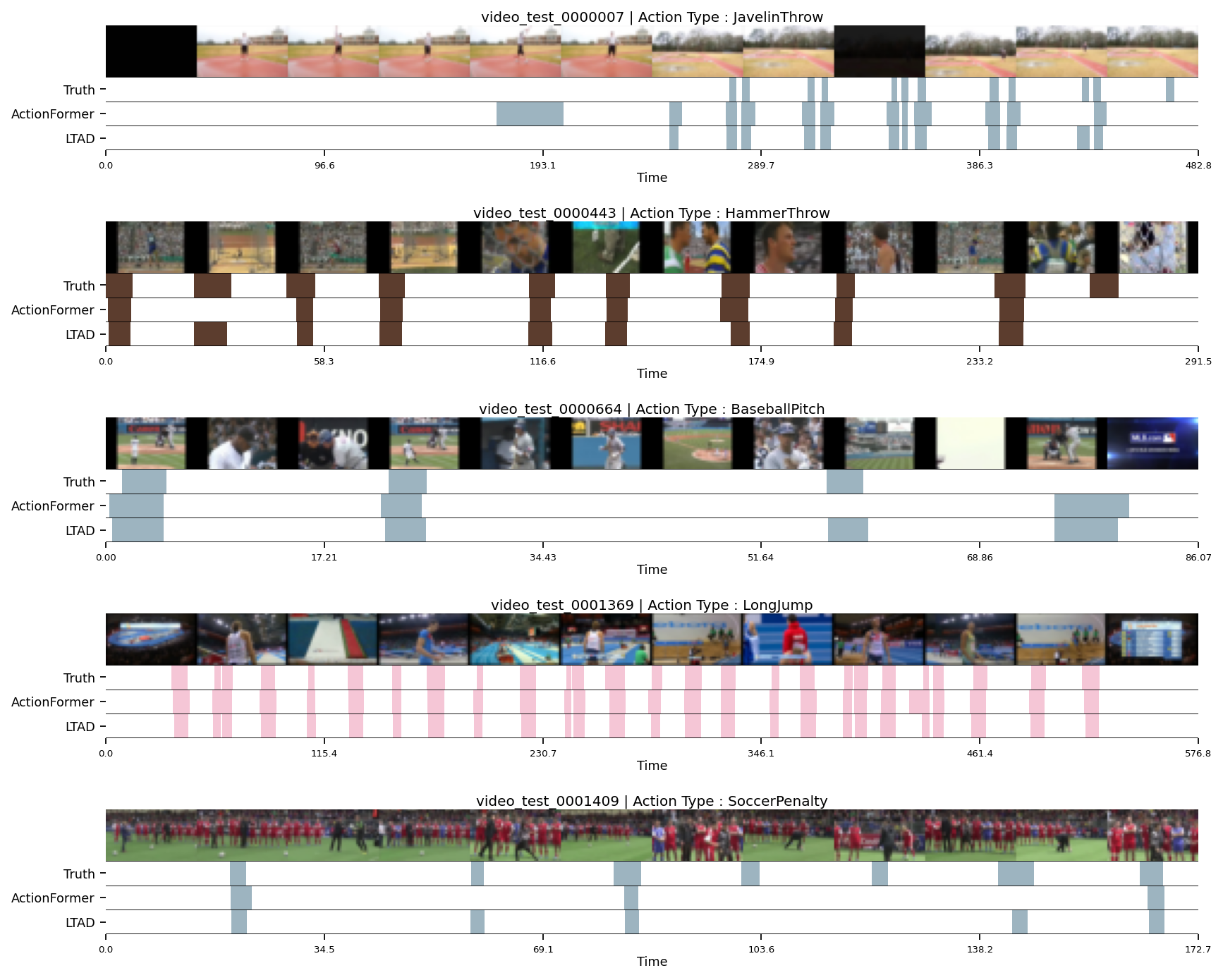}
  \caption{Qualitative visualization of action detection results. We compare the predicted action boundaries of LiquidTAD (LTAD) against the strong baseline ActionFormer~\cite{zhang2022actionformer} and the Ground Truth (Truth). Across various action types (e.g., JavelinThrow, HammerThrow, BaseballPitch), LiquidTAD produces more coherent temporal boundaries and reduces fragmented predictions.}
  \label{fig:qualitative}
\end{figure}

Figure~\ref{fig:qualitative} provides qualitative comparisons between LiquidTAD, ActionFormer~\cite{zhang2022actionformer}, and ground truth annotations on THUMOS-14. Across several action types, LiquidTAD produces more coherent temporal boundaries and generates fewer fragmented predictions within continuous action instances. These observations are consistent with the quantitative results and suggest that the proposed relaxation-based temporal modeling helps reduce fragmented predictions.

\subsection{Hardware-Agnostic Deployment and Latency}
A primary limitation of SSM-based models such as MambaTAD~\cite{mambatad2024} is their reliance on hardware-specific custom CUDA kernels, which restricts deployment flexibility. To evaluate LiquidTAD's hardware-agnostic efficiency, we conduct a CPU inference benchmark across all methods using the same input tensor specification. All latency results are measured on the temporal detection head only, using a synthetic input tensor of shape $(1, 2304, C)$ with batch size 1, where $C$ denotes the channel dimension of each model's default configuration. Experiments are conducted on an Intel Xeon Gold 6240 CPU (2.60GHz) using PyTorch 2.5.1, with single-thread inference and no post-processing (NMS excluded). Feature extraction backbones are excluded from all measurements.

As shown in Table~\ref{tab:latency}, TriDet~\cite{shi2023tridet} and ActionFormer~\cite{zhang2022actionformer} record CPU latencies of 1512ms and 1516ms, respectively. LiquidTAD achieves a latency of 782ms, approximately half that of these baselines. Although TemporalMaxer~\cite{tang2023temporalmaxer} achieves lower CPU latency (615ms), LiquidTAD obtains a higher average mAP on THUMOS-14 while remaining substantially more efficient than attention-based and SSM-based detectors. MambaTAD~\cite{mambatad2024} incurs a latency of over 57,400ms on the same hardware---approximately $73\times$ higher than LiquidTAD---suggesting that its selective scan implementation is less favorable on CPUs when optimized GPU kernels are unavailable. LiquidTAD is built entirely on standard PyTorch operators and thus requires no hardware-specific fallback. To assess robustness, we re-evaluate the same checkpoint under CPU precision on THUMOS-14, observing a drop of 0.59 percentage points in average mAP (from 69.46\% to 68.87\%).

% CPU 延迟对比表
\begin{table}[htbp]
  \centering
  \caption{CPU inference latency comparison ($T=2304$, batch size 1, single thread, detection head only). Models relying on custom CUDA kernels incur substantial latency increases on general-purpose hardware.}
  \renewcommand{\arraystretch}{1.1}
  \resizebox{\columnwidth}{!}{
    \begin{tabular}{l c cc}
    \toprule
    \textbf{Method} & \textbf{Latency (ms)} $\downarrow$ & \textbf{GFLOPs} & \textbf{Params (M)} \\
    \midrule
    TemporalMaxer~\cite{tang2023temporalmaxer} & 615.13  & 23.55 & 7.12  \\
    \textbf{LiquidTAD}                        & \textbf{782.08}  & 27.17 & 10.82 \\
    TriDet~\cite{shi2023tridet}                & 1512.09 & 43.84 & 15.99 \\
    ActionFormer~\cite{zhang2022actionformer}  & 1516.20 & 45.41 & 29.25 \\
    DyFADet~\cite{dyfahdet2024}                & 3662.32 & 91.07 & 27.59 \\
    VideoMambaSuite~\cite{chen2024videomambasuite}    & 10327.53& 43.22 & 20.34 \\
    MambaTAD~\cite{mambatad2024}           & 57458.62& -     & 27.77 \\
    \bottomrule
    \end{tabular}%
  }
  \label{tab:latency}
\end{table}

\subsection{Ablation Studies}

\textbf{1) Parallel Liquid-inspired Relaxation Backend.}
To demonstrate the advantage of our parallel formulation, we compare it against sequential liquid-inspired backends in Table~\ref{tab:training_speed}. The ODE~\cite{chen2018neuralode} and CfC~\cite{hasani2022cfc} backends require 3663.83s and 1683.08s per training epoch, respectively, due to their recurrent state update procedures. In contrast, the proposed parallel relaxation backend reduces training time to 12.96s per epoch while achieving the highest average mAP of 69.46\% among the compared backends. This efficiency gain stems from the non-recursive formulation: by fixing the decay factor as a block-wise scalar and eliminating hidden state propagation, the temporal update becomes a fully vectorized operation with no sequential dependency. Training and inference times are measured on the full THUMOS-14 validation set using a single GPU.

The accuracy advantage of the parallel backend over the dynamic ODE and CfC formulations~\cite{hasani2021ltc,hasani2022cfc} is also expected rather than incidental. In standard CfC/LTC networks, the input-dependent time constant $\tau(x_t, u_t)$ is recomputed at each step based on the current network state. However, in the TAD setting, input features are pre-extracted by powerful off-the-shelf backbones (e.g., I3D~\cite{carreira2017i3d}) that already encode rich local and global temporal context. Consequently, the additional representational capacity offered by a dynamic $\tau$ provides limited benefit, while its sequential computation introduces substantial optimization overhead over long sequences. The static, per-block decay rate $\lambda_l$, by contrast, provides a stable and scale-aligned inductive bias better suited to this pre-extracted feature regime, confirming that the parallel design is an empirically motivated architectural choice rather than a mere computational simplification.

% 训练效率对比表
\begin{table}[htbp]
  \centering
  \caption{Training and inference efficiency of different liquid-inspired relaxation backends on THUMOS-14.}
  \renewcommand{\arraystretch}{1.1}
  \resizebox{\columnwidth}{!}{
    \begin{tabular}{l c cc}
    \toprule
    \textbf{Backend} & \textbf{Avg mAP (\%)} & \textbf{Train Time / Ep. (s)} & \textbf{Total Inference Time (s)} \\
    \midrule
    ODE~\cite{chen2018neuralode,hasani2021ltc} & 68.23 & 3663.83 & 4595.28 \\
    CfC~\cite{hasani2022cfc}                   & 68.41 & 1683.08 & 2674.00 \\
    \textbf{Parallel (Ours)} & \textbf{69.46} & \textbf{12.96} & \textbf{65.00} \\
    \bottomrule
    \end{tabular}
  }
  \label{tab:training_speed}
\end{table}

\textbf{2) Decay-Rate Parameterization Strategy.}
Table~\ref{tab:tau_ablation} validates the design of our Hierarchical Decay-Rate Sharing Strategy. Constraining the decay rate to a block-wise shared scalar improves average mAP from 68.62\% (per-channel decay rate) to 69.46\%. This confirms that aligning the temporal decay rate with the scale-specific properties of the feature pyramid~\cite{lin2017fpn} effectively mitigates optimization noise and avoids introducing redundant channel-level decay parameters.

% Decay-rate 消融表
\begin{table}[htbp]
  \centering
  \caption{Ablation on decay-rate parameterization strategy on THUMOS-14.}
  \renewcommand{\arraystretch}{1.1}
  \resizebox{\columnwidth}{!}{
    \begin{tabular}{l ccccc c}
    \toprule
    \textbf{Decay-Rate Strategy} & \textbf{0.3} & \textbf{0.4} & \textbf{0.5} & \textbf{0.6} & \textbf{0.7} & \textbf{Avg mAP} \\
    \midrule
    Per-channel decay rate & 83.17 & 79.13 & 72.33 & 61.48 & 46.98 & 68.62 \\
    \textbf{Block-wise shared decay rate} & \textbf{84.21} & \textbf{79.86} & \textbf{72.90} & \textbf{62.87} & \textbf{47.47} & \textbf{69.46} \\
    \bottomrule
    \end{tabular}%
  }
  \label{tab:tau_ablation}
\end{table}

\textbf{3) Ablation on Discrete Time Step ($\Delta t$).}
Table~\ref{tab:dt_ablation} reports the effect of the structural time step $\Delta t$ on detection accuracy. The default value $\Delta t = 4/30$ corresponds to the temporal stride of the pre-extracted I3D~\cite{carreira2017i3d} features under a 30 FPS assumption, and yields the best average mAP of 69.46\%. Notably, scaling $\Delta t$ to align with the downsampling strides of each pyramid level (\texttt{align\_dt\_pyramid}) leads to a substantial drop to 65.44\%.

We attribute this to the temporal compression inherent in deep feature pyramids. While early-stage features closely correspond to physical video time, successive downsampling causes deeper tokens to represent aggregated temporal windows rather than discrete physical timestamps. Directly scaling $\Delta t$ according to pyramid strides may therefore over-constrain high-level semantic features whose temporal correspondence has already been compressed by downsampling. This observation further motivates our Hierarchical Decay-Rate Sharing Strategy: rather than imposing a fixed physical time prior via $\Delta t$, LiquidTAD learns layer-adaptive decay rates that implicitly compensate for pyramid-level temporal compression.

% Delta t 对比表
\begin{table}[htbp]
  \centering
  \caption{Ablation on the structural time step $\Delta t$ on THUMOS-14.}
  \renewcommand{\arraystretch}{1.1}
  \resizebox{0.85\columnwidth}{!}{
    \begin{tabular}{l c}
    \toprule
    \textbf{Time Step Configuration ($\Delta t$)} & \textbf{Avg mAP (\%)} \\
    \midrule
    $2/30$ & 68.23 \\
    $\mathbf{4/30}$ (Default) & \textbf{69.46} \\
    $4/30$ + \texttt{align\_dt\_pyramid} & 65.44 \\
    $8/30$ & 68.21 \\
    $1.0$ & 67.43 \\
    \bottomrule
    \end{tabular}%
  }
  \label{tab:dt_ablation}
\end{table}

\textbf{4) Feature Pyramid Depth.}
Table~\ref{tab:pyramid_depth} ablates the number of pyramid levels. A 6-level configuration achieves the best average mAP of 69.46\%. Shallower pyramids (4-level: 66.72\%) lack sufficient temporal scale coverage for longer actions, while deeper pyramids (7- and 8-level: 68.50\% and 68.61\%) introduce excessive temporal compression at higher levels, degrading detection of shorter actions. The 6-level structure provides the best trade-off across action durations.

\begin{table}[htbp]
  \centering
  \caption{Ablation on feature pyramid depth on THUMOS-14.}
  \renewcommand{\arraystretch}{1.1}
  \resizebox{0.6\columnwidth}{!}{
    \begin{tabular}{l c}
    \toprule
    \textbf{Pyramid Levels} & \textbf{Avg mAP (\%)} \\
    \midrule
    4-level & 66.72 \\
    5-level & 68.41 \\
    \textbf{6-level (Default)} & \textbf{69.46} \\
    7-level & 68.50 \\
    8-level & 68.61 \\
    \bottomrule
    \end{tabular}%
  }
  \label{tab:pyramid_depth}
\end{table}

\section{Conclusion}

In this paper, we presented \textbf{LiquidTAD}, an efficient TAD framework that distills the exponential relaxation prior of liquid neural dynamics~\cite{hasani2021ltc,hasani2022cfc} into a non-recursive parallel temporal operator. By combining Parallel Liquid-inspired Relaxation with a Hierarchical Decay-Rate Sharing Strategy, LiquidTAD captures scale-aware temporal dynamics using only standard neural operations, avoiding sequential ODE solving~\cite{chen2018neuralode} and specialized kernels. Experiments on THUMOS-14~\cite{idrees2017thumos} and ActivityNet-1.3~\cite{caba2015activitynet} show that LiquidTAD achieves competitive detection accuracy with substantially reduced model complexity, obtaining 69.46\% average mAP on THUMOS-14 with only 10.82M parameters and 27.17G FLOPs. These results demonstrate that liquid-inspired temporal relaxation provides a lightweight and deployment-friendly alternative for efficient temporal action detection.

% ===================== REFERENCES =====================

\end{document}